\pdfoutput=1
\documentclass{llncs}
\usepackage{llncsdoc}    
\usepackage[utf8]{inputenc}
\begin{document}
\newcounter{save}\setcounter{save}{\value{section}}
{\def\addtocontents#1#2{}%
\def\addcontentsline#1#2#3{}%
\def\markboth#1#2{}%
\title{A Corpus of English-Hindi Code-Mixed Tweets for Sarcasm Detection}

\author{\textbf{Sahil Swami}, \textbf{Ankush Khandelwal}, \textbf{Vinay Singh}, \textbf{Syed Sarfaraz Akhtar} \and \textbf{Manish Shrivastava}}
\institute{Language Technologies Research Centre, International Institute of Information Technology, Hyderabad}

\maketitle
\begin{abstract}
Social media platforms like twitter and facebook have become two of the largest mediums used by people to express their views towards different topics. Generation of such large user data has made NLP tasks like sentiment analysis and opinion mining much more important. Using sarcasm in texts on social media has become a popular trend lately. Using sarcasm reverses the meaning and polarity of what is implied by the text which poses challenge for many NLP tasks. The task of sarcasm detection in text is gaining more and more importance for both commercial and security services. We present the first English-Hindi code-mixed dataset of tweets marked for presence of sarcasm and irony where each token is also annotated with a language tag. We present a baseline supervised classification system developed using the same dataset which achieves an average F-score of 78.4 after using random forest classifier and performing 10-fold cross validation.
\end{abstract}
\section{Introduction}
The Oxford dictionary\footnote{http://www.oxforddictionaries.com/} defines sarcasm as:
``the use of irony to mock or convey contempt". Sarcasm generally has an implied negative statement but a positive surface sentiment \cite{1}. As an example, consider the tweet: ``\textit{I'm so happy the teacher gave me all this homework right before Spring Break}". The author of this tweet uses positive words like `happy' but it can be clearly observed that the author is not happy. Although sarcasm cannot be completely formally defined, it can be detected by humans in texts and speech. Sarcasm and irony,though different, are very closely related \cite{2}, so we consider them same in this paper.
Twitter is one of the most used social media platforms used by people to express their opinion \cite{3}. Many companies use this data for opinion mining and sentiment analysis to study the market. But a tweet may not always state the exact opinion of the user i.e. if it is sarcastically expressed. As it has become a common trend to use sarcasm on social media texts, detecting sarcasm in a tweet becomes more crucial for tasks like opinion mining and sentiment analysis.

Code-switching and code-mixing are two of the most commonly studied phenomena in multilingual societies \cite{4}. Code-switching is generally inter-sentential while code-mixing is intra-sentential. Code-mixing refers to embedding linguistic units of one language into an utterance of another language. An example of a code-mixed sentence is: ``\textit{modi ji notebandi ki dikkat ko door karne k liye 200 rupay ka note bi market me laao. Chae 50 ka band ho jaye}". Words such as `market' are in English, and words like `ki', `door', etc. are Hindi words which are transliterated to English.
Hindi is the most spoken language in India and fourth most spoken in the world whereas English is the third most spoken language in the world. Thus there are a lot of people who express themselves on social media using English-Hindi code-mixed texts which makes sarcasm detection in such texts much more important.

Several experiments of sarcasm detection have been performed on tweets in English \cite{5},\cite{2},\cite{11} as well as in other languages such as Czech \cite{6}, Dutch \cite{7} and Italian \cite{8} but there have been no experiments on English-Hindi code-mixed texts mainly because of the lack of annotated resources.

The main contribution of this paper is to provide a resource of English-Hindi code-mixed tweets which contain both sarcastic and non-sarcastic tweets. We provide tweet level annotation for presence of sarcasm and token level language annotation. This corpus can be used to train, develop and also evaluate the performances of sarcasm detection and language identification techniques on a code-mixed corpus. In addition, we present a baseline supervised classification system for sarcasm detection developed using the same corpus.

Both the dataset and classification system are available online\footnote{https://github.com/sahilswami96/SarcasmDetection\_CodeMixed}.
\section{Dataset}

\subsection{Data Collection}
To collect sarcastic tweets we extract tweets containing hashtags \#sarcasm and \#irony \cite{9} 
using the Twitter Scraper API and manually select English-Hindi code-mixed tweets from them. We also use other keywords such as `bollywood', `cricket' and `politics' to collect sarcastic tweets from these domain. Out of these collected tweets, sarcastic and non-sarcastic tweets are further manually separated. To collect more non-sarcastic tweets we extract tweets with keywords such as `bollywood', `cricket' and `politics' which do not contain hashtags \#sarcasm and \#irony. Further English-Hindi code-mixed tweets are manually selected from them. Having only sarcastic or only non-sarcastic tweets from a particular domain may lead to an unbiased classification system therefore we make sure that there are both sarcastic and non-sarcastic tweets from each domain. The twitter scraper API collects each tweet in json format after which we extract the tweet content and tweet id from it. Figure 1. shows an example of a tweet collected in json format.
\subsection{Data Processing and Annotation}
Tweets are annotated by a group of people fluent in both English and Hindi. Each tweet is manually annotated for presence of sarcasm. Tweets are then tokenized and each token is annotated with a language which is manually verified. We used Cohen’s Kappa [16] as a measure of inter-annotator agreement and it was calculated to be 0.79.
\subsubsection{Sarcasm Annotation}
Each tweet is manually annotated for presence of sarcasm using the tags `YES' and `NO'. Tweets with the hashtags \#sarcasm and \#irony are more likely to contain sarcasm. Tweets which do not contain these hashtags are then manually verified to not contain sarcasm. An example of a tweet (with translation in English) that contains sarcasm and one that does not:\\

\textit{Tweet: @bonda0123 sir g .. \#insomniac likhte ho aur jaldi sone ki baat bhi karte ho !! \#irony !!}\\
\textit{Translation: @bonda0123 sir You write \#insomniac and talk about sleeping early !! \#irony !!}\\
Sarcasm: YES\\

\textit{Tweet: Bhai kuchh bhi karna iss @SimplySajidK ke saath movie mat karna..Bollywood se nafrat ho jaati hai..Itni sadi hui ghatiya filmein banata h ye}
\textit{Translation: Brother do anything but don't do a movie with @SimplySajidk..I start hating Bollywood..They make such bad films}\\
Sarcasm: NO\\

Hashtags \#sarcasm and \#irony are randomly removed from some tweets which contain sarcasm so that the dataset contains both types of sarcastic and ironic tweets, ones with the hashtags \#sarcasm and \#irony and ones without.
\subsubsection{Tokenization and Language Annotation}
There have been several experiments of language identification \cite{10},\cite{13} on various types of texts which motivates the task of token level language annotation in this dataset.
Each tweet is tokenized using white spaces as delimiters and taking into account the trends found in the dataset such as use of multiple consecutive punctuations, mentions, etc. Each token is annotated with a language tag. One of the following tags is assigned for language: `en', `hi' and `rest', where `en' stands for English, `hi' for Hindi and `rest' for punctuations, emoticons, named entities, URLs, etc. 'en' is assigned to English words such as `play', `warm', etc. and `hi' is assigned to Hindi words transliterated in English such as `sahi', 'kya'. Initially each token is assigned language tags using online dictionaries such as Enchant and the `rest' tags are assigned by identifying hashtags, URLs and mentions. Every language tag and token is manually verified to correct any mistakes in language tags and tokenization. An example of a tweet with language tags:

\begin{table}[h!]
\begin{center}
\begin{tabular}{|c|c|}

\hline
\textbf{Token} & \textbf{Language}\\
\hline
bhai&hi\\
\hline
triple&en\\
\hline
talaq&hi\\
\hline
se&hi\\
\hline
aap&hi\\
\hline
kya&hi\\
\hline
samjhte&hi\\
\hline
hai&hi\\
\hline
samjhaye&hi\\
\hline
aap&hi\\
\hline
zara&hi\\
\hline
..&rest\\
\hline
agar&hi\\
\hline
triple&en\\
\hline
talaq&hi\\
\hline
pta&hi\\
\hline
hota&hi\\
\hline
apko&hi\\
\hline
toh&hi\\
\hline
aisa&hi\\
\hline
nhi&en\\
\hline
kehte&hi\\
\hline
..&rest\\
\hline
\end{tabular}
\caption{A tweet with token level language annotation}
 \end{center}
\end{table}
\subsection{Dataset analysis}
The dataset consists of 5250 English-Hindi code-mixed tweets out of which 504 tweets are marked as sarcastic and ironic. The dataset consists of two types of tweets:
1.) Tweets that are marked as sarcastic but do not have hashtags \#sarcasm or \#irony present in them.
2.) Tweets that contain these hashtags but are not marked as sarcastic.
This sparsity in the corpus also helps in developing a better system for sarcasm detection.
The average length of a tweet is 22.2 tokens per tweet. The average number of tokens per tweet annotated with `en', `hi' and `rest' tags are 2.1, 16.1 and 4.0 respectively. Table 2. and Table 3. show corpus level and tweet level statistics respectively.
\begin{table}[h!]
\centering
\begin{tabular}{|c|c|}
\hline
\textbf{Category} & \textbf{Number of tweets}\\
\hline
Total tweets&5250 \\
\hline
Sarcastic tweets&504 \\
\hline
Non-sarcastic tweets&4746 \\
\hline
 \end{tabular}
\caption{Corpus level statistics}
\end{table}
\begin{table}[h!]
\centering
\begin{tabular}{|c|c|}
\hline
\textbf{Category} & \textbf{Number of tokens}\\
\hline
Avg. tokens&22.2 \\
\hline
Avg. en tokens&2.1 \\
\hline
Avg. hi tokens&16.1 \\
\hline
Avg. rest tokens&4.0  \\
 \hline
 \end{tabular}
\caption{Tweet level statistics}
\end{table}
\subsection{Dataset structure}
The corpus is structured into three files. The first file contains a tweet id followed by the corresponding tweet text and a blank line and so on. The second file consists of tweet ids followed by language annotated tweets as depicted in Table 1. The third file has the annotation for presence of sarcasm for each tweet. Each tweet id is followed by one of the sarcasm label, a blank line.

\section{Sarcasm detection system}
We present a baseline classification system for sarcasm detection in English-Hindi code-mixed tweets using various word based and character based features. We run and compare various machine learning models which use these features to detect sarcasm.

\subsection{Preprocessing}
It is a common practice on social media to use camel case while writing hashtags. Thus we extract the hashtags from each tweet and extract separate tokens from it by removing the `\#' and using a hashtag decomposition approach \cite{16} assuming it is written in camel case. For example we can get `I', `Am' and `Sarcastic' from `\#IAmSarcastic'. URLs, mentions, stop words and punctuations are removed from tweets for further processing.
\subsection{Features}
\subsubsection{Word N-Grams}
Word n-gram refers to presence or absence of contiguous sequence of n word or tokens in a tweet. Word n-grams have proven to be useful features for sarcasm detection in previous experiments \cite{11},\cite{2},\cite{6}. We consider all n-grams for values of n ranging from 1 to 5. We consider only those n-grams for features which occur at least 10 times in the corpus in order to prune the feature space.
\subsubsection{Character N-Grams}
Character n-gram refers to presence or absence of contiguous sequence of n characters in a tweet. It can be observed from previous experiments \cite{2},\cite{6} that character n-grams play an important role in sarcasm detection. We consider all n-grams for values of n ranging from 1 to 3. If we include all these character n-grams then it will increase the size of the feature vectors enormously thus we consider only those n-grams which occur at least 8 times in the dataset.
\subsubsection{Sarcasm Indicative Tokens}
This feature refers to the presence or absence of sarcasm indicative tokens. We use a variation of the approach \cite{14} to find indicative hashtags and extract sarcasm indicative tokens for each language label. We calculate a score for each token where score is defined as:

\[Score(token) =
max_{label\in{Sarcasm-Set}} \frac{freq(token,sarcasm\_label)}{freq(token)}\]
where Sarcasm-Set = \{YES, NO\}.

We consider only those tokens as features for sarcasm indication which have a score $\geq$ 0.6 and occur at least five times in the dataset. We find such tokens for each of the language tags and consider them in the feature vector. The threshold value for scores and number of occurrences has been decided after empirical fine tuning.
\subsubsection{Emoticons}
This feature refers to the presence or absence of various emoticons in the tweet. There have been several experiments \cite{5},\cite{6} where emoticons are used as a feature for sarcasm detection. We consider a set of 27 emoticons as features.  
\subsection{Feature Selection}

It has been observed in various experiments \cite{12},\cite{15} that feature selection algorithms improve the performance of machine learning models significantly. We use chi square feature selection algorithm which uses chi-squared statistic to evaluate individual feature with respect to each class. This algorithm was used in order to extract the best features and reduce the feature vector size to 500.
\subsection{Classification Approach}
We use three classification techniques:  Support Vector Machine with Radial Basis Function kernel, Linear Support Vector Machine, and Random Forest classifier. We use scikit-learn implementation of these methods for sarcasm detection. We also perform 10-fold cross validation on the corpus created to develop the system. 10-fold cross validation is run for each of the individual features separately to observe the effect of each feature on classification.
\subsection{Results}
We use F-score measure to evaluate the performance of our system as the number of sarcastic tweets is less than the number of non-sarcastic tweets and thus using just accuracy for evaluation of the system may not be a good metric. F-score is defined as the harmonic mean of precision and recall. 

\[F-score=2\frac{precision*recall}{precision+recall}\]\\

Precision and recall are defined as:

\[Precision=\frac{tp}{tp+fp}\]

\[Recall=\frac{tp}{tp+fn}\]

where \textit{tp}, \textit{fp} and \textit{fn} are true positives, false positives and false negatives respectively.

Our system achieves a best average F-score of 78.4 after running 10-fold cross validation using the random forest classifier on the dataset.
Table 2. shows the F-scores achieved by each of the systems for each feature separately as well as with all the features combined.
It can be observed that each feature affects each technique differently. Word n-grams perform best with random forest classifier whereas character n-grams with RBF kernel SVM and sarcasm indicative tokens perform best with linear svm.

\begin{table}[h!]
\centering
\begin{tabular}{|c|c|c|c|}
\hline
\textbf{Features} & \textbf{RBF Kernel SVM} & \textbf{Random Forest} & \textbf{Linear SVM}\\
\hline
Character n-grams&73.1&75.0&66.4 \\
\hline
Word n-grams&71.4&76.7&68.0 \\
\hline
Sarcasm indicative tokens&66.1&72.0&70.2 \\
\hline
Emoticons&62.8&68.5&65.7 \\
\hline
All features&76.5&78.4&71.7  \\
 \hline
 \end{tabular}
\caption{F-scores for all the three classifiers}
\end{table}

\section{Conclusion}
With the increase in number of people using social media to express their views, tasks like opinion mining and sentiment analysis have gained a lot of importance. And using sarcasm in these social media texts make these tasks much more challenging.

We presented the first English-Hindi code-mixed dataset for sarcasm detection collected from twitter. We explained the methods used for collecting and annotating these tweets at both tweet level for presence of sarcasm as well as at token level for language. We also presented a baseline supervised classification developed using the same dataset which uses three different machine learning techniques and 10-fold cross validation.

\section{Future work}
This dataset can further be normalized at token level which will thus improve the performance of the classification system. This dataset can also be used to develop systems for automatic language identification in code-mixed texts.

Similar datasets can be created for different language pairs with the presene of other emotions such as humor.

The provided classification system can be improved further by using various other features such as word embeddings, POS tags and other language based features.


\end{document}